\pdfoutput=1
\documentclass[11pt]{article}

\usepackage{ACL2023}

\usepackage{times}
\usepackage{latexsym}
\usepackage{booktabs}
\usepackage{graphicx}
\usepackage{amsmath}
\usepackage{amssymb}
\usepackage[T1]{fontenc}
\usepackage[utf8]{inputenc}

\usepackage{microtype}

\usepackage{inconsolata}
\usepackage{pifont}
\usepackage{soul}
\usepackage{xcolor}

\definecolor{DarkGreen}{HTML}{388E3C}

\newcommand{\redcross}{\textcolor{red}{\ding{55}}} % Red cross
\newcommand{\greencheck}{\textcolor{DarkGreen}{\ding{51}}} % Green greencheck

\title{\raisebox{-0.1\height}{\includegraphics[height=1.5em]{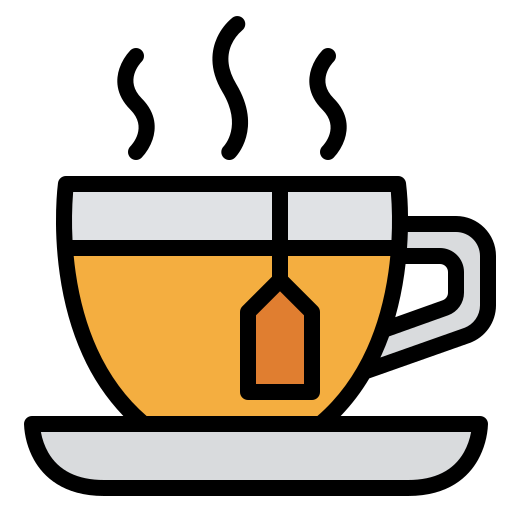}}
  ELAICHI: Enhancing Low-resource TTS by Addressing Infrequent and Low-frequency Character Bigrams
}

\author{Srija Anand*,  Praveen Srinivasa Varadhan*, Mehak Singal, Mitesh M. Khapra \\
  Nilekani Centre at AI4Bharat, Indian Institute of Technology Madras, India \\
  \texttt{\{da24s013, cs21d201\}@smail.iitm.ac.in} \\
  \href{https://github.com/AI4Bharat/ELAICHI}{\raisebox{-0.2\height}{\includegraphics[height=1.2em]{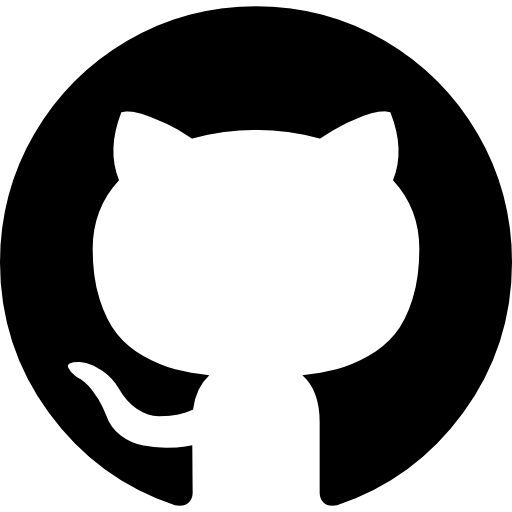}} ELAICHI}
  }

\begin{document}
\maketitle
\begin{abstract}
Recent advancements in Text-to-Speech (TTS) technology have led to natural-sounding speech for English, primarily due to the availability of large-scale, high-quality web data. However, many other languages lack access to such resources, relying instead on limited studio-quality data. This scarcity results in synthesized speech that often suffers from intelligibility issues, particularly with low-frequency character bigrams. In this paper, we propose three solutions to address this challenge. First, we leverage high-quality data from linguistically or geographically related languages to improve TTS for the target language. Second, we utilize low-quality Automatic Speech Recognition (ASR) data recorded in non-studio environments, which is refined using denoising and speech enhancement models. Third, we apply knowledge distillation from large-scale models using synthetic data to generate more robust outputs. Our experiments with Hindi demonstrate significant reductions in intelligibility issues, as validated by human evaluators. We propose this methodology as a viable alternative for languages with limited access to high-quality data, enabling them to collectively benefit from shared resources. 
\end{abstract}

\section{Introduction}

In recent years, substantial progress has been made in text-to-speech (TTS) synthesis for English, with state-of-the-art models generating highly natural speech \cite{shen2024naturalspeech, ju2024naturalspeech}. %, peng2024voicecraft}. 
Unlike traditional TTS systems, which relied on limited studio-quality data, modern models are trained on web-scale datasets from diverse sources %— comprising podcasts, audiobooks, movies, and other diverse sources — that
spanning various emotions and speaking styles \cite{lyth2024NaturalLG}. %This advancement in English is largely enabled by 
The ready availability of such data coupled with the development of advanced automatic speech recognition (ASR) systems \cite{radford2023whisper, zhang2023usm, pratap2020mls}, which facilitate accurate transcription, results in rich audio-text pairs for training.

In contrast, low-resource languages lack both extensive web-scale datasets \cite{doddapaneni-etal-2023-towards} and accurate ASR systems. As a result, TTS systems for these languages lag significantly behind, often relying on small, curated datasets 
\cite{Kumar2022Towards, prakash2022exploring, prakash2023towards}.
Due to the limited size of the datasets and the prevalence of code-mixing which introduces English origin words, various character bigrams are not adequately represented in these datasets. Consequently, these systems often exhibit intelligibility issues, particularly on test utterances containing such low-frequency character bigrams. To overcome these limitations, we propose three methods in this work aimed at improving intelligibility for low-frequency character bigrams.    

First, we investigate the use of additional studio-quality data from related low-resource languages. This study focuses on Hindi, one of the most widely spoken languages in the Indian subcontinent, with approximately 528.3 million speakers globally. Many languages in this region have shared vocabulary, highly overlapping phoneme sets and compatible orthographic systems with similar grapheme-to-phoneme mappings. 

Consequently, jointly training a model on data from multiple such languages may improve the representation of low-frequency character bigrams, improving performance in the target language. In this work, we explore the possibility of using additional data from (i) one related language which uses the same script, (ii) multiple languages with overlapping phoneme sets but not necessarily using the same script, and (iii) a variety of speakers across related languages, rather than relying on a single male or female speaker per language.     

Second, while studio-quality data may be limited, many low-resource languages possess low-quality audio-text pairs initially used for training ASR systems. We propose repurposing such data for training TTS systems by applying denoising \cite{liu2021voicefixer} and enhancement techniques \cite{schroeter2023deepfilternet3}. Specifically, we leverage the IndicVoices-R dataset \cite{sankar2024indicvoices-r}, where denoising and speech enhancement models were employed to clean an existing ASR dataset \cite{javed2024indicvoices}, resulting in higher-quality data suitable for TTS training. A key feature of this dataset is its inclusion of spontaneous speech on diverse topics from a demographically varied group of speakers. Additionally, it captures colloquial and code-mixed language nuances, which may improve the representation of low-frequency bigrams in current TTS datasets, thereby enhancing the  intelligibility and naturalness of synthesized speech.

Third, we draw inspiration from recent advancements in distilling knowledge from large-scale models to enhance the performance of smaller models \cite{gandhi2023distilwhisper, abdin2024phi, dubey2024llama, chevi2023nix}. Specifically, we utilize the pre-trained VoiceCraft checkpoint from IndicVoices-R \cite{sankar2024indicvoices-r}, which was trained on enhanced ASR data from multiple Indian languages. Using this model, we synthesize speech samples for utterances that contain low-frequency bigrams.
By augmenting the existing high-quality TTS dataset with synthetic speech targeting these low-frequency character bigrams, we aim to improve their coverage and, consequently, the intelligibility of TTS systems.
Our experiments with Hindi as the target language show that all our proposed methods outperform the baseline. We explain these improvements using corpus-level statistics, specifically (i) \textit{coverage} and (ii) \textit{relative growth} of low-frequency character bigrams. Multilingual training yielded the highest intelligibility rates due to significant \textit{relative growth}. It also achieved high mean opinion scores, showing the effectiveness of scaling with studio-quality datasets from multiple languages. Lastly, our experiments with synthetic data reveal that simply increasing the \textit{coverage} of low-frequency bigrams is not sufficient; ensuring adequate representation of each new bigram in the augmented corpus is also crucial. 
We hope our findings can serve as a foundation for improving TTS systems in other low-resource languages.
\section{Related Work}

\textbf{State of English TTS:}
Intelligibility is a critical aspect of Text-to-Speech (TTS) systems \cite{tan2021survey}. 
Several works \cite{shen2024naturalspeech, ju2024naturalspeech, cassanova2024xtts, peng2024voicecraft} have achieved state-of-the-art performance in English, excelling in both naturalness and intelligibility. 
Additionally, assessing performance on out-of-distribution (OOD) texts has gained traction, with methods like StyleTTS \cite{li2023styletts} enhancing the generalization capabilities of TTS systems for OOD content. However, a key enabler of these advances in English TTS is the massive availability of high-quality data \cite{pratap2020mls, chen2021gigaspeech}, which low-resource languages often lack. Thus, several strategies that are well-motivated and necessary for low-resource languages become less relevant for English, where there is data abundance.

\noindent \textbf{Low-Resource TTS Intelligibility: } 
Limited training data in low-resource languages often results in unnatural and difficult-to-understand speech. Prior work \cite{anand2024enhancing} has shown that TTS systems for Indian languages struggle with missing character bigrams, and propose a low-cost solution through volunteer-recorded data to improve performance. In contrast, our work expands the scope by demonstrating that Hindi TTS systems also exhibit poor synthesis on low-frequency character bigrams. We address this issue by proposing methods that use related language data, ASR datasets, or outputs from large pre-trained models, thus reducing the need for manual data collection.

\noindent \textbf{Augmentation Strategies:}
Several augmentation strategies have been explored to improve TTS systems. Multilingual pre-training \cite{amalas2024multilingual} has significantly improved the naturalness and intelligibility of TTS systems for low-resource languages compared to monolingual models. This work investigates multilingual training's effectiveness in addressing pronunciation errors related to low-frequency character bigrams. Another established approach is augmenting TTS datasets with synthetic data; for example, \cite{lajszczak2022} showed that synthetic audio samples created by substituting text and audio fragments improve expressive TTS quality. Similarly, IndicVoices-R \cite{sankar2024indicvoices-r} demonstrated that scaling TTS models with ASR-enhanced data boosts few-shot and zero-shot speaker performance. Our work builds on these strategies to improve intelligibility of low-frequency character bigrams.

\section{Background}

Let $\mathcal{C}$ be a speech corpus defined as a set of $N$ utterances, denoted as $\mathcal{C} = \{u_1, u_2, \ldots, u_N\}$. Each utterance $u_i$ is represented as a text-audio pair $u_i = (t_i, a_i)$,
where $t_i$ is the text uttered and $a_i$ is the corresponding audio recording. Every text utterance $t_i$ is a sequence of characters, $c_j^i$ such that $t_i = [c_1^i, c_2^i, \dots, c_{L_i}^i]$, where $L_i$ is the length of the $i^{th}$ text utterance.

\noindent \textbf{Definition } \textit{(Character bigram)}. We consider a \textit{character bigram} as a pair of consecutive characters in an utterance. More formally, a character bigram for the $i$-th utterance is defined as:
\[
b_j^i = (c_j^i, c_{j+1}^i),
\]
for $1 \leq j \leq L_{i - 1}$. The set of all character bigrams from the entire corpus is denoted as $\mathcal{B}$:
\[
\mathcal{B} = \bigcup_{i=1}^{N} \{b_1^i, b_2^i, \dots, b_{L_{i-1}}^i\}.
\]

\noindent \textbf{Definition } \textit{(Frequency)}. For each character bigram $b \in \mathcal{B}$, its frequency is defined as the number of occurrences across all utterances in the corpus:
\[
f(b) = \sum_{i=1}^{N} \sum_{j=1}^{L_{i-1}} \mathbb{I}(b_j^i = b),
\]
where $\mathbb{I}(\cdot)$ is the indicator function that returns 1 if the character bigram $b_j^i = b$ and 0 otherwise.

\noindent \textbf{Definition } \textit{(Low Frequency Character bigram)}.
A character bigram is considered low-frequency (LF) if:
\[
f(b) < t
\] where $t \in \mathbb{Z}^+$ is a frequency threshold. The threshold $t$ can be selected based on various criteria such as quantiles of the frequency distribution, standard deviations or the Pareto principle. In this work, we initialize $\mathcal{L}$ from the bigram set $\mathcal{B}_{\mathcal{T}}$ of a larger text corpus $\mathcal{T}$ to ensure broad coverage of bigrams and to accurately capture infrequent patterns. Leveraging a large corpus allows us to model the full distribution of character bigrams across diverse contexts, ensuring that our identification of low-frequency character bigrams is robust and not biased by the smaller training data. We set the threshold $t = 40$, corresponding to the 40th percentile of the frequency distribution.
%}

\noindent The set of low-frequency character bigrams $\mathcal{L}$ is then defined as:
\[
\mathcal{L} = \{ b \in \mathcal{B}_{\mathcal{T}} \mid f(b) < t \}.
\]
Our objective is to show that TTS models tend to produce higher error rates for text sequences containing character bigrams from $\mathcal{L}$. We hypothesize that low-frequency character bigrams correspond to phonetic patterns that the model has not adequately learned due to insufficient training examples. 
%\comment{Below to end section new, please review.}
To mitigate this issue, we propose methods that augment the original corpus and increase the frequency of bigrams $\in \mathcal{L}$. Let the augmented corpus be $\Hat{\mathcal{C}}$. We define two new measures to quantitatively reflect the improvement of the augmented corpus over the original one in terms of low-frequency character bigrams. 

\noindent  \textbf{Definition} \textit{(Coverage)} measures the absolute increase in the number of low-frequency bigrams introduced in the augmented corpus compared to the original corpus. We define it as,
\[
\Delta F = \left| \Hat{\mathcal{B}} \cap \mathcal{L} \right| - \left| \mathcal{B} \cap \mathcal{L} \right|
\]
where $\Hat{\mathcal{B}}$ is the set of all character bigrams in $\Hat{\mathcal{C}} $.

\noindent \textbf{Definition} \textit{(Relative Growth)} measures the change in frequency of LF character bigrams across the original and augmented corpora, accounting for both existing and new bigrams. We define the growth in low-frequency bigrams for $\Hat{\mathcal{C}}$ as, 
\[
\Delta G = \sum_{b \in \Hat{\mathcal{B}} \cap \mathcal{L}} \Hat{f}(b) - \sum_{b \in B \cap \mathcal{L}} f(b)
\]
where, $\Hat{\mathcal{B}}$ is the set of bigrams in $\Hat{\mathcal{C}}$ characterized by frequency function $\Hat{f}$. The relative growth is given by, 

\[
\Delta G_{\text{rel}} = \frac{\Delta G}{\sum_{b \in B \cap \mathcal{L}} f(b)} 
\]      

\section{Methodology}

We propose four methods to mitigate the errors on low-frequency bigrams by augmenting the training data with additional sources. The objective is to improve the representation of rare bigrams through carefully designed data augmentation strategies. 
Let the training corpus, set of bigrams, frequency, and set of low-frequency bigrams of the training corpus of the baseline method be given by $\mathcal{C}$, $\mathcal{B}$, $f$, and, $\mathcal{L}$, respectively.

\subsection{Adding a Proximal Language}
\label{method_proximal_lang}

In this method, we augment our corpus by integrating character bigrams from a related language that shares the same script as the target language. Let \(\mathcal{P}\) denote the proximal language speech corpus.  The inclusion of the proximal language introduces additional character bigrams into our training dataset.

%\add{
Let $\mathcal{B_\mathcal{P}}$ represent the set of bigrams of $\mathcal{P}$. 
%}
The augmented corpus \(\Hat{ \mathcal{C} }\) is defined as:
\[
\Hat{ \mathcal{C} } = \mathcal{C} \cup \mathcal{P}.
\]
The set of bigrams in the enhanced corpus is then given by $\Hat{\mathcal{B}} = \mathcal{B} \cup \mathcal{B}_{\mathcal{P}}.$ As a result of this augmentation, we realize that the frequency of low-frequency bigrams may increase. Specifically, for any bigram \(b \in \Hat{\mathcal{B}}\), its updated frequency \(\Hat{f}(b)\) can be expressed as:
\[
\Hat{f}(b) = f(b) + \sum_{i=1}^{N_\mathcal{P}} \sum_{i=1}^{L_i^\mathcal{P}-1} \mathbb{I}(b_j^i = b),
\]
where $b^i_j$ is the $j^{th}$ bigram in the $i^{th}$ utterance of $\mathcal{P}$, $L^\mathcal{P}_j$ is the length of the $i^{th}$ text utterance, and $N_\mathcal{P}$ is the total number of utterances in $\mathcal{P}$. Consequently, some bigrams in the original low-frequency set \(\mathcal{L}\) may increase in frequency and potentially rise above the threshold \(t\). 

%\add{
\[
\Delta G = \sum_{b \in \Hat{\mathcal{B}} \cap \mathcal{L}} \sum_{j=1}^{N_\mathcal{P}} \sum_{k=1}^{L_j^{\mathcal{P}}-1} \mathbb{I}(b_j^i = b)
\]
%}
This method seems promising as it utilizes overlapping bigrams from proximal languages, and we hypothesize that, by doing so, it is able to enhance the model's ability to generalize and accurately pronounce previously underrepresented low-frequency bigrams.

\subsection{Scaling Multilingual Data}
\label{method_multi_lang}

In this method, we enhance our training corpus by incorporating data from multiple languages, including both those that share the same script as the target language and those with different scripts. Let \(\mathcal{M}\) denote the multilingual speech corpus consisting of utterances from various languages. The augmented corpus \(\Hat{ \mathcal{C} }\) is defined as $\Hat{ \mathcal{C} } = \mathcal{C} \cup \mathcal{M}.$ Likewise, the set of character bigrams for the enhanced corpus is given by: $\Hat{\mathcal{B}} = \mathcal{B} \cup \mathcal{B}_{\mathcal{M}}.$
Incorporating data from multiple languages not only increases the frequency counts of shared bigrams for languages with the same script but may also allow the model to learn implicit acoustic representations for overlapping phoneme bigrams in languages with different scripts. 
%\add{
The growth in $\Hat{\mathcal{C}}$ is given by, 
\[
\Delta G = \sum_{b \in \Hat{\mathcal{B}} \cap \mathcal{L}} \sum_{l \in \Gamma} \sum_{i=1}^{N_l} \sum_{j=1}^{L_i^{l}-1} \mathbb{I}(b_j^{i,l} = b)
\]
where $\Gamma$ represents the set of languages sharing the same script as the target language, $N_l$ is the number of utterances in the $l^{th}$ language, and $L_i^{l}$ is the length of the $i^{th}$ utterance in the $l^{th}$ language.
%}

\subsection{Adding ASR-Enhanced Data}

Automatic Speech Recognition (ASR) data is a widely available resource that typically introduces a large number of speakers into the training set. While ASR datasets can help increase the frequency counts of low-frequency bigrams in the target language, they often come with lower-quality transcripts and suboptimal audio recordings. These factors can potentially degrade the performance of the TTS system. To mitigate this, we employ denoised and dereverberated ASR data in our experiments to ensure better audio quality.

The methodology for augmenting the corpus with ASR data follows a process similar to that of adding a proximal language, but with the crucial difference that here we add ASR data from the same language. The augmented corpus \(\Hat{\mathcal{C}}\) is defined as $\Hat{\mathcal{C}} = \mathcal{C} \cup \mathcal{A}$, 
where \(\mathcal{A}\) represents the ASR corpus in the target language. The set of bigrams from the ASR-enhanced corpus is then given by \(\Hat{\mathcal{B}} = \mathcal{B} \cup \mathcal{B}_{\mathcal{A}}\), where $\mathcal{B}_{\mathcal{A}}$ represents the set of bigrams of $\mathcal{A}$.

%\add{
\noindent The growth in $\Hat{\mathcal{C}}$ is given by, 
\[
\Delta G = \sum_{b \in \Hat{\mathcal{B}} \cap \mathcal{L}} \sum_{i=1}^{N_\mathcal{A}} \sum_{j=1}^{L_i^\mathcal{A}-1} \mathbb{I}(b_j^i = b)
\]
where \(N_\mathcal{A}\) is the number of utterances in the ASR corpus, and \(L_i^\mathcal{A}\) is the length of the \(i^{th}\) utterance.
%}

\subsection{Adding Synthetic Data}
\label{synth_data}
In this method, we generate synthetic speech data by distilling outputs from a large, pre-trained TTS model that has been trained on a much larger corpus. This large model would typically perform well on a variety of low-frequency character bigrams due to its exposure to diverse data (typically, from multiple languages) and greater capacity. The large capacity and corresponding large size of this model would make it impractical for deployment in downstream use cases. However, by using this model to generate synthetic data, we can augment the original corpus and improve the performance of a smaller, more efficient TTS model. This approach is practical because the smaller model can then be trained on this enriched dataset, benefiting from the knowledge of the larger model while maintaining computational efficiency during inference.

Let $\mathcal{S}$ be the synthetic corpus, wherein we ensure that every utterance covers at least one low-frequency bigram $b \in \mathcal{L}$. The augmented corpus \(\Hat{\mathcal{C}}\) is then defined as $\Hat{\mathcal{C}} = \mathcal{C} \cup \mathcal{S}$.

% \add{
The growth in $\Hat{\mathcal{C}}$ is given by, 
\[
\Delta G = \sum_{b \in \Hat{\mathcal{B}} \cap \mathcal{L}}  \sum_{i=1}^{N_\mathcal{S}} \sum_{j=1}^{L_i^\mathcal{S}-1} \mathbb{I}(b_j^i = b)
\]
where \(N_\mathcal{S}\) is the number of utterances in $\mathcal{S}$, and \(L_i^\mathcal{S}\) is the length of the \(i^{th}\) utterance.
%}

\section{Experimental Setup}
In this section we outline the datasets and models used across our experiments and detail the evaluation setup of our experiments. Table \ref{tab:data_stats} reports the data statistics for our proposed methods.
\begin{figure}[!t]

    \centering
    \includegraphics[width=0.47\textwidth]{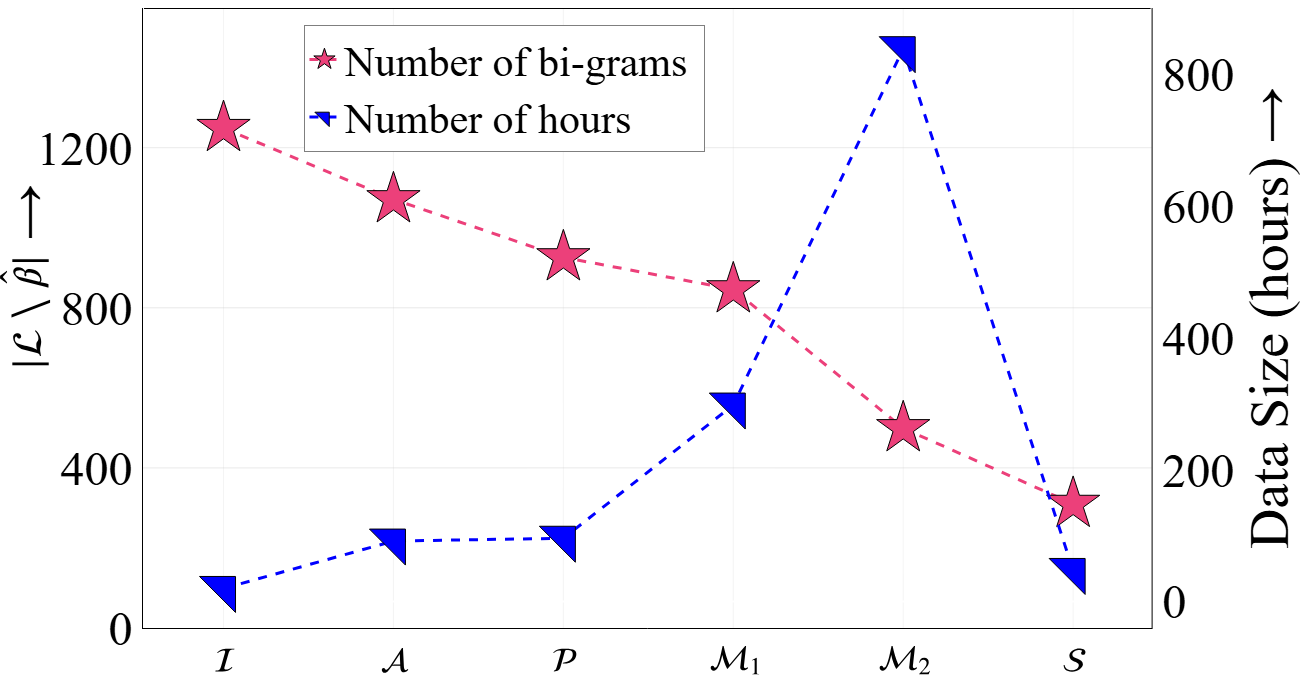}

    \caption{A comparison of the number of bigrams from $\mathcal{L}$ missing in the corpora (left Y axis) and the training data size (right Y axis) of the corpora under consideration.}
    \label{fig:bigram-vs-hours} 
\end{figure}

\subsection{Datasets}
\label{datasets}

\begin{table}[]
\fontsize{10pt}{10pt}\selectfont
\centering
\begingroup
\setlength{\tabcolsep}{6pt} 
\centering

\begin{tabular}{@{}lccrrrr@{}}
\toprule
\textbf{Corpus}           & \textbf{Studio} & \textbf{Duration (h)} & \textbf{\# Spk} & \textbf{\# Lang} \\ \midrule
$\quad \mathcal{I}$              & \greencheck & 20.16         & 2             & 1             \\
$\quad \mathcal{P}$              & \greencheck       & 80.11         & 2             & 1             \\
$\quad \mathcal{A}$              & \redcross        & 72.74         & \textbf{368}           & 1             \\
$\;\; \mathcal{M}_1$            & \greencheck       & 224.57        & 27            & 14            \\
$\;\; \mathcal{M}_2$            & \greencheck      & \textbf{852.17} & 260          & \textbf{15}            \\ 
$\quad \mathcal{S}$              & \redcross       & 27.67         & 1             & 1             \\ \bottomrule
\end{tabular}
\endgroup
\caption{Statistics of corpora used in different experimental setups, where \# Spk indicates number of speakers, and \# Lang indicates number of languages.} 
\label{tab:data_stats}
\end{table}
\noindent \textbf{Massive Text Corpus ($\mathcal{T}$)}:  We collate a substantial text corpus using resources such as Sangraha \cite{khan2024indicllmsuite}, the Bharat Parallel Corpus \cite{gala2023indictrans2}, and transcriptions from IndicVoices \cite{javed2024indicvoices}. We define the low-frequency bigram set $\mathcal{L}$ using this corpus by setting the threshold, $t=40$. 

\noindent\textbf{Baseline Corpus ($\mathcal{I}$)}: All our experiments include the baseline dataset which is 20 hours of Hindi from IndicTTS \cite{baby2016resources}. This set contains 200 low-frequency bigrams.

\noindent \textbf{Proximal Language Corpus ($\mathcal{P}$)}: Chhattisgarhi is an Eastern dialect of Hindi and shares the Devanagari script with Hindi. Therefore, we initialize $\mathcal{P}$ with 80 hours of high-quality, studio-recorded Chhattisgarhi data from LIMMITS \cite{limmits2024}.

\noindent \textbf{Multilingual Corpus ($\mathcal{M}$)}: We conduct two experiments to investigate the effect of adding multilingual data, one at a smaller scale and one by pooling most publicly available Indian TTS datasets. In the first experiment, we initialize $\mathcal{M}_1$ with the IndicTTS dataset~\cite{baby2016resources}, which includes 13 official languages of India - Assamese, Bengali, Gujarati, Kannada, Malayalam, Manipuri, Odia, Punjabi, Tamil, Telugu, Bodo, Hindi, and Marathi, of which the latter three use Devanagri script. Each language has approximately 20 hours of high-quality, studio-recorded data, evenly split between a female and a male speaker. We include $\mathcal{P}$ in $\mathcal{M}_1$, thus the dataset contains 14 languages.

\noindent In the second experiment, we scale up to include multiple TTS datasets for Indian languages. $\mathcal{M}_2$ includes IndicTTS \cite{baby2016resources}, LIMMITS \cite{limmits2024}, Google Crowdsourced \cite{butryna2020google}, and Rasa \cite{ai4bharat2024rasa}, totaling approximately 852 hours of speech. We thus extend our dataset to include 15 Indian languages, with addition of Nepali, and a total of 260 speakers. In Figure~\ref{fig:bigram-vs-hours}, we see that this method adds the most number of hours and effectively scales down the number of missing low-frequency character bigrams to 499.

\noindent \textbf{ASR Corpus ($\mathcal{A}$)}: We include the 72 hours of the Hindi subset of IndicVoices-R \cite{sankar2024indicvoices-r} in $\mathcal{A}$. The dataset includes both read and extempore speech, spanning over 21 domains, as outlined in \cite{javed2024indicvoices}, providing a comprehensive coverage of character bigrams.

\noindent \textbf{Synthetic Corpus ($\mathcal{S}$)}: We see in Figure \ref{fig:bigram-vs-hours} that incorporating various corpora, such as proximal languages and ASR increases the data size but still insufficiently covers the low-frequency bigrams. To address this, we derive a subset of $\mathcal{T}$ by greedily selecting sentences containing low-frequency bigrams, as discussed in Section~\ref{synth_data}. We then synthesize these utterances, resulting in a synthetic Hindi speech corpus, $\mathcal{S}$. Figure~\ref{fig:bigram-vs-hours} illustrates that just by adding 28 hours of synthetic audio, we are left with only 310 missing low-frequency bigrams which is the least amongst all methods. We discuss the model used to generate $\mathcal{S}$ in Section \ref{subsec: models}. 

\subsection{Models}
\label{subsec: models}

\noindent \textbf{VITS: }  The VITS \cite{kim2021conditional} model used in all our experiments consists of a posterior encoder, prior encoder, decoder, discriminator, and stochastic duration predictor. The posterior encoder, which is only used during training, utilizes non-causal WaveNet \cite{oord2016wavenet} residual blocks to model the posterior distribution, with global conditioning for speaker embeddings in multi-speaker scenarios. The prior encoder includes a Transformer-based \cite{vaswani2017attention} text encoder with relative positional encoding, followed by a normalizing flow based on WaveNet residual blocks for flexible prior distribution modeling. The decoder adopts HiFi-GAN V1 \cite{kong2020hifi}, leveraging transposed convolutions and multi-receptive field fusion for waveform generation. The discriminator employs a multi-period architecture, and the stochastic duration predictor estimates phoneme durations using dilated convolutions and neural spline flows. We use default hyperparameters and train on 8 NVIDIA A100 GPUs.

\noindent \textbf{Synthetic Data Generation: }
We employ VoiceCraft \cite{peng2024voicecraft} to generate the synthetic corpus as it is trained on a large dataset, with diverse speakers and accents, and it excels in producing natural, intelligible speech with minimal errors. VoiceCraft is a state-of-the-art neural codec language model that employs a Transformer decoder architecture, incorporating a token rearrangement procedure that leverages causal masking and delayed stacking to generate speech. Specifically, we leverage the pre-trained checkpoint of VoiceCraft from IndicVoices-R \cite{sankar2024indicvoices-r}, trained on over 1704 hours of Indian language data.

\subsection{Evaluation}

We primarily rely on human intelligibility tests to evaluate whether TTS performance has improved for low-frequency character bigrams. Additionally, we rely on Mean-Opinion-Scores (MOS) and use automated measures for speech-quality assessment to ensure that the overall quality of TTS has not degraded due to the augmentation of training data from various sources.

\noindent \textbf{Intelligibility Rates (IR): } We task human raters to evaluate whether the low-frequency bigrams in the TTS output are intelligible. Each rater listens to an audio sample and views the corresponding text with the low-frequency bigrams highlighted. They are asked to give a binary rating indicating whether the specific segment of speech corresponding to the bigram is intelligible. Here, raters are instructed to focus solely on the intelligibility of the marked segment, ignoring issues in other parts of the utterance. The test interface is implemented using Label Studio \cite{labelstudio}, where raters can control the playback speed. To ensure clarity, all audios are initially played at 0.5x speed, and raters can adjust as needed. Two native speakers of the language rated 100 utterances each for each method.

\noindent \textbf{Mean Opinion Scores (MOS): } While the focus of the above test was only on the performance on low-frequency bigrams, we also do a MOS test to gauge whether the improved performance on low-frequency bigrams leads to any deterioration in the overall speech quality. In this test, listeners were asked to rate the quality of the audio samples on a scale of 1 to 5, with 1 being ``Poor'' and 5 being ``Excellent''. Listeners were instructed to focus on several key aspects of the speech: overall intelligibility, pronunciation, and naturalness (as opposed to focusing only on low-frequency character bigrams).

\noindent \textbf{Automated Speech-Quality Assessment: } We use Brouhaha Speech-to-Noise-Ratio (SNR) \cite{lavechin2023brouhaha} as a measure of speech levels to noisy levels. We also measure reverbrant effects using C50 scores from the same implementation. Since the addition of numerous speakers in the augmented corpus can deteriorate the speaker resemblance of the target speaker of the baseline model, we also assess speaker similarity by comparing the synthesized outputs to the original speaker's voice. Particularly, we calculate cosine similarity between the embeddings of ground-truth and synthesized audio, which are extracted using Titanet \cite{koluguri2022titanet}. These automated measures ensure that the quality of TTS has not degraded due to the augmentation of the original corpus to enhance low-frequency character bigram data. 

\section{Results and Discussion}
In this section, we analyze the improvements of TTS systems trained with our proposed methods compared to the baseline, with a primary focus on intelligibility rates (IR), as presented in Table \ref{main-results}. To explain these IR scores, we examine two key factors: \textit{relative growth} ($\Delta G_{\text{rel}}$) and \textit{coverage} ($\Delta F$). Additionally, we assess the speech quality of each method using MOS scores and automated speech evaluation metrics, as shown in Table \ref{tab:auto-speech-eval}. 

\begin{table*}[t!]

\centering
\begin{tabular}{@{}llrrcc@{}}
\toprule
\textbf{Method} &
  \textbf{Corpus} &
  \multicolumn{1}{c}{\textbf{$\Delta F$}} &
  \multicolumn{1}{c}{\textbf{$\Delta G_{\text{rel}}$}} &
  \textbf{IR} &
  \textbf{MOS} \\ \midrule
Baseline             & I      & -  & \multicolumn{1}{c}{-} & 0.51                        & 3.97                        \\
Proximal             & I + $\mathcal{P}$   & 326  & 35.1                  & 0.74                        & 4.16                        \\
ASR-Enhanced         & I + $\mathcal{A}$   & 178  & 1.42                  & 0.66                        & 3.85                        \\
Multilingual         & I + $\mathcal{M}_1$  & 403  & 36.51                 & 0.85                        & 4.42                        \\
Scaling Multilingual & I + $\mathcal{M}_2$  & 750  & \textbf{129.73}                & \textbf{0.90}               & \textbf{4.51}               \\
Synthetic            & I + $\mathcal{S}$   & \textbf{939} & 7.92                  & 0.71                        & 4.08                        \\ \bottomrule
\end{tabular}%

\caption{A comparison of the proposed methods - with the 
The intelligibility Error Rates for Low Frequency bigrams (IR) and MOS scores for the methods proposed.}
\label{main-results}

\end{table*}

\subsection{Effect of Using a Proximal Language}
In Table \ref{main-results}, we see that adding the proximal language Chhattisgarhi ($\mathcal{P}$) for Hindi brings the \textit{coverage} ($\Delta F$) to 326 and improves IR from 0.51 to 0.74, a 45\% increase. This improvement is potentially guided by the high \textit{relative growth} ($\Delta G_{\text{rel}} = 35.1$), indicating significant increase in the frequency of low-frequency (LF) bigrams. We see that the addition of more data also improves the MOS of the TTS system.

\subsection{Effect of Using Restored ASR Data}
Although adding the restored ASR data ($\mathcal{A}$) increases the \textit{coverage} to 178, it remains lower than that achieved by $\mathcal{P}$. The \textit{relative growth} is also minimal, at just 1.42, reflecting a smaller increase in the frequency of LF bigrams compared to other methods. As expected, this modest improvement in \textit{coverage} translates into a smaller gain in intelligibility rates, as reflected by an IR score of 0.66. It is worth noting that the ASR-enhanced dataset is comparable in size to the proximal dataset, making the difference in performance more evident. Finally, the MOS score sees a slight reduction compared to the baseline, dropping to 3.85, suggesting that the lower quality of ASR data slightly affects the naturalness of the synthesized speech.

\subsection{Effect of Multilingual Augmentation}
The use of multilingual augmentation ($\mathcal{M}1$) significantly enhances the TTS system's performance compared to the proximal language method ($\mathcal{P}$). Specifically, \textit{coverage} ($\Delta F$) improves by 23.6\%, increasing from 326 to 403, while maintaining a comparable \textit{relative growth} ($\Delta G_{\text{rel}}=36.51$) to that of the proximal method. In terms of intelligibility rates (IR), the system shows a remarkable 67\% improvement over the baseline, with an IR of 0.85, and a further 15\% increase compared to the proximal setup. Additionally, the benefits of scaling the multilingual data are evident in the MOS scores, which rise to 4.42, further indicating an enhancement in speech quality alongside the gains in intelligibility. %We refer to this model as ELAICHI and will publicly release it.

\subsection{Impact of Scaling Languages and Speakers}
While it may seem intuitive that increasing the dataset size improves performance, we emphasize that the key factor driving the improvement in intelligibility rates (IR) is the augmentation with corpora exhibiting high values of \textit{coverage} and \textit{relative growth}. This can be seen in Table \ref{main-results}, where the scaled multilingual dataset ($\mathcal{M}_2$) achieves 1.86x higher coverage than $\mathcal{M}_1$ and 3.55x greater relative growth, resulting in an IR of 0.90—a significant 76.47\% improvement over the baseline. Although the dataset size scales by 628 hours compared to $\mathcal{M}_1$, the corresponding improvement in IR is relatively modest at 5.88\%. Nevertheless, this scaling leads to the highest MOS score of 4.51, highlighting that while intelligibility sees marginal gains, speech quality benefits substantially from the additional data. 

\subsection{Effect of addition of synthetic data}
Given the substantial computational costs involved in generating speech data with large models, we prioritize maximizing \textit{coverage} over \textit{relative growth} within the synthetic corpus. Specifically, our approach emphasizes the inclusion of utterances that introduce new low-frequency (LF) bigrams into the dataset, rather than producing multiple utterances for the same LF bigrams.
This strategy results in an improvement in intelligibility rates (IR) relative to the baseline, however, the IR remains lower than those achieved with $\mathcal{M}_1$, $\mathcal{M}_2$, and $\mathcal{P}$. Notably, despite achieving the highest $\Delta F$ of 939, the IR does not reach the performance levels of these other methods. This observation underscores the necessity for both high \textit{coverage} and substantial \textit{relative growth} to drive significant improvements in IR.
Interestingly, while the gains in intelligibility are modest, the mean opinion scores (MOS) do not decline, indicating that the synthetic data generation model successfully preserves speech quality comparable to the baseline. Given our limited computational budget, we could not do an additional experiment where we improve \textit{relative growth} in addition to \textit{coverage}. We leave this as future work.

\begin{table}[t]
% \centering
\fontsize{10pt}{10pt}\selectfont
\centering
\begingroup
\setlength{\tabcolsep}{6pt} % Default value: 6pt
\begin{tabular}{@{}lccc@{}}
\toprule
\textbf{Method}           & \textbf{S-SIM} & \textbf{SNR}   & \textbf{C50}   \\ \midrule
Baseline                  & 0.73           & 67.03          & 59.60          \\
Proximal              & \textbf{0.73}  & 67.63          & \textbf{59.72}          \\
ASR-Enhanced          &  0.72  & 67.50          & 59.62          \\
Multilingual         & 0.67           & 66.81          & 59.67          \\
Scaling Multilingual & 0.72           & \textbf{68.04} & 59.71 \\
Synthetic            & 0.72           & 67.27          & 59.53          \\ \bottomrule
\end{tabular}
\caption{Automated speech evaluation metrics for baseline and proposed methods.}
\label{tab:auto-speech-eval}
\endgroup
\end{table}

\subsection{Automated Speech Quality Assessment}
We evaluate the quality of the synthesized speech of the target speaker of the baseline TTS system, which is susceptible to change after augmenting the training corpus with the methods discussed before. We evaluate on three automated metrics: speaker similarity (S-SIM), speech-to-noise ratio (SNR), and speech clarity (C50). The results for each method are summarized in Table~\ref{tab:auto-speech-eval}, showing that the proposed methods maintain speech quality.

\noindent \textbf{Speaker Similarity: } The S-SIM scores indicate that the speaker similarity remains fairly consistent across methods. The Proximal methods achieves the highest speaker similarity scores (0.73), while Multilingual ($\mathcal{M}_1$), show slight decreases in speaker similarity. This may be due to the fact that Proximal and ASR-Enhanced methods add more speakers from the same or closely related languages, preserving speaker characteristics, whereas Multilingual scaling introduces a richer variety of speakers,  including those from other language families, which could contribute to the slight degradation in speaker identity. Despite the variation, all methods maintain speaker consistency, indicating that the augmentation of training data does not lead to significant degradation.

\noindent \textbf{Speech-To-Noise Ratio: } The SNR scores reveal that all methods effectively maintain a high level of speech clarity against background noise, with the ($\mathcal{M}_2$) method achieving the highest score of 68.04. Overall, these results indicate that augmenting the training corpus does not adversely affect the SNR of synthesized speech compared to the baseline.

\noindent \textbf{Speech Clarity: } The C50 scores, which measure the reverberation in speech, are consistent across the proposed methods, all hovering around 59.60 to 59.72. However, we expect the ASR-Enhanced and Synthetic methods to show lower C50 scores, as all other methods introduce studio-quality data. Despite this expectation, the ASR-Enhanced method scores 59.62, which reflects the effectiveness of speech-enhancement on noisy data. Conversely, the Synthetic method displays a slight dip in C50 from the baseline, which could be attributed to a cascade effect where a large model (VoiceCraft) is trained on ASR-enhanced data (IndicVoices-R), then synthesized ($\mathcal{S}$), and subsequently used to train VITS on the synthesized outputs. Overall, the results affirm that nearly all the proposed methods effectively preserve speech clarity, thus supporting their viability for practical deployment.

\section{Conclusion}
In this work, we tackle the challenge of improving intelligibility for low-frequency character bigrams in TTS systems. Addressing the lack of large web-scale datasets for low resource languages, we propose three methods for leveraging existing speech corpora: incorporating proximal languages, augmenting with ASR-enhanced data, and generating synthetic data. Our experiments demonstrate that each method improves the intelligibility rates of these low-frequency character bigrams compared to the baseline with the multilingual scaling of data yields the best intelligibility and MOS scores due to significant coverage and relative growth. %Notably, our careful curation of synthetic data performs better than ASR-enhanced corpus with one-third the data size.
Further our analysis reveals that coverage alone is insufficient, proper representation of newly introduced bigrams is crucial. These results suggest that resource-constrained languages can significantly enhance TTS systems by carefully augmenting their data with related languages and synthetic approaches. All our models will be publicly released.

\section*{Limitations}

Although we maximize \textit{coverage} of low-frequency bigrams during synthetic generation, we are unable to maximize \textit{relative growth}. This was mainly due to the computational costs of generation using a large 880M parameter model as a result of which we had to limit the size of the synthetic corpus to 28 hours. If these barriers can be transgressed, by scaling up synthetic generation on a much larger corpus, the benefits of synthetic generation may be better realized.

Additionally, in this work we focus only on Hindi, as it allowed us to do an in-depth analysis. While we believed these findings would be relevant to other Indian languages, going forward, compute constraints not withstanding, we would like to focus on other languages that may have different phonetic structures or cultural contexts that could influence TTS performance.

\section*{Ethics Statement}
We prioritized ethical conduct throughout all stages of this research. In the evaluation of our models, we engaged expert listeners with prior experience in assessing TTS systems. These raters, all native Hindi speakers with the required language proficiency, were essential in providing accurate and meaningful feedback on the intelligibility and naturalness of the synthesized speech. We ensured that all raters were compensated fairly in accordance with industry standards for their time and expertise. In addition, raters were fully informed about the purpose of the experiments, their role, and their right to withdraw from the study at any point without any consequences, ensuring their autonomy and informed consent throughout the process.

We use open-sourced code for our trainings and automatic speech assessments. We only use ChatGPT for assistance purely with the language of the paper, i.e., paraphrasing, spell-checking, or polishing the author’s original content, without suggesting new content.

We release all our model checkpoints under the CC BY 4.0 license. While there is potential for misuse, we believe that an open-source model for a low-resource language is essential not only for advancing research but also for facilitating practical applications and deployment in real-world settings. By making our model openly available, we aim to empower communities working on underrepresented languages to build more natural-sounding TTS systems, thereby fostering linguistic diversity and inclusion in technology. Our release is intended to support ethical use cases, such as education, accessibility, and preserving endangered languages, and we encourage responsible use of the technology.

\bibliography{custom}
\bibliographystyle{acl_natbib}

\end{document}